\documentclass[10pt,twocolumn,letterpaper]{article}

\usepackage{wacv}              % To produce the CAMERA-READY version
\definecolor{wacvblue}{rgb}{0.21,0.49,0.74}
\usepackage[pagebackref,breaklinks,colorlinks,allcolors=wacvblue]{hyperref}
\usepackage{adjustbox}
\usepackage{multirow}
\usepackage{booktabs}
\usepackage{tabularx}
\usepackage{array}

\usepackage{listings}
\def\wacvPaperID{2323} % *** Enter the WACV Paper ID here
\def\confName{WACV}
\def\confYear{2027}

\title{Learning Fine-Grained Visual Classification with Text as Privileged Information via Multimodal Distillation}

\author{
  Aishwarya Venkataramanan \quad Christian Ickler \quad Joachim Denzler \\
  Computer Vision Group, Friedrich Schiller University Jena \\
  \ttfamily aishwarya.venkataramanan@uni-jena.de
}

\begin{document}
\maketitle
\begin{abstract}
Fine-grained visual classification (FGVC) requires distinguishing between highly similar categories by focusing on subtle, localized visual differences. Recent multimodal methods have demonstrated that textual attribute descriptions can provide valuable semantic supervision by explicitly describing discriminative object characteristics. However, these descriptions are typically unavailable at test time, limiting their direct use in deployment.
We propose FG-LUPI, a Learning Using Privileged Information (LUPI)-based framework that leverages attribute descriptions during training. We train a multimodal teacher that integrates image patches and textual attributes via gated cross-attention, allowing semantic attributes to condition visual representations.
We distill this knowledge into an image-only student using multi-level distillation at the prediction, patch, and attention levels, aligning global predictions, patch representations, and attention patterns with the teacher. This encourages the student to inherit attention structure learned from multimodal supervision while maintaining image-only inference.
Experiments on multiple FGVC benchmarks show consistent improvements in classification accuracy over strong ViT-based baselines. Furthermore, evaluation using Expected Calibration Error (ECE) shows that FG-LUPI also produces better-calibrated predictions across all six benchmarks. These results suggest that textual attribute supervision can benefit not only fine-grained recognition performance but also the reliability of model confidence, while preserving the efficiency and practicality of image-only inference.
\end{abstract}
    
\section{Introduction}
\label{sec:intro}

Fine-grained visual classification (FGVC) aims to distinguish between highly similar categories by identifying subtle visual differences related to appearance, texture, shape, and object parts. This task is challenging due to strong inter-class similarity and intra-class variations~\cite{venkataramanan2021tackling, venkataramanan2023gaussian}. As a result, FGVC models must effectively localize and exploit discriminative visual features.

%\begin{figure}[t]
%    \centering
%    \includegraphics[width=0.75\linewidth]{images/fig_1.pdf}
%    \caption{\textbf{Conceptual overview of FG-LUPI.} During training (top), a multimodal teacher learns from both images and privileged textual descriptions, allowing textual attributes to influence the learned visual representations. Through multi-level distillation, an image-only student learns to mimic the teacher's predictions, representations, and attention patterns. During inference (bottom), the student performs classification using only images.}

%    \label{fig:lupi_illustration}
%\end{figure}

\begin{figure}[t]
\centering
\setlength{\tabcolsep}{1pt} % or 0pt

\begin{tabular}{@{}ccc@{}}
Original & Vanilla ViT & FG-LUPI \\

\includegraphics[width=0.3\linewidth]{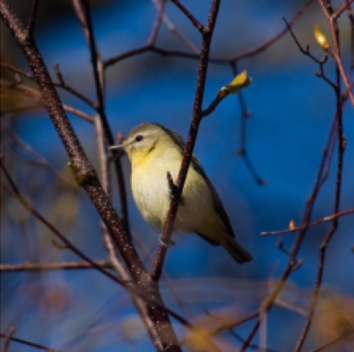} &
\includegraphics[width=0.3\linewidth]{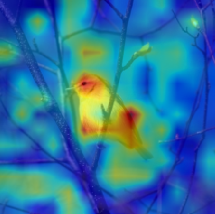} &
\includegraphics[width=0.3\linewidth]{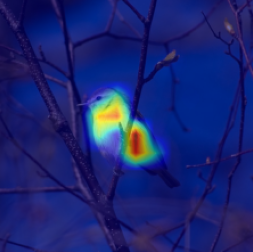} \\

\includegraphics[width=0.3\linewidth]{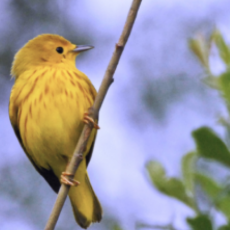} &
\includegraphics[width=0.3\linewidth]{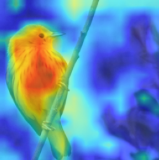} &
\includegraphics[width=0.3\linewidth]{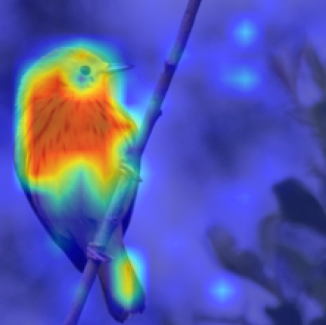} \\

\end{tabular}

\caption{\textbf{Qualitative comparison of attention maps.} Attention map visualization on CUB-200-2011 for a vanilla ViT and the proposed FG-LUPI student model. Compared to the vanilla model, FG-LUPI exhibits more concentrated attention on the bird and reduced attention to surrounding background regions.}

\label{fig:vit_vs_ours}
\end{figure}

To address this challenge, prior work has explored part-based modeling~\cite{han2019p,korsch2019classification,korsch2021end}, attention mechanisms~\cite{he2022transfg,wang2021feature}, and language-guided supervision for fine-grained recognition~\cite{akataEvaluationOutputEmbeddings2015, jiang2024delving, xiao2025flair, venkataramanan2023integrating}. Natural language provides complementary semantic information that explicitly describes discriminative attributes. Such descriptions often highlight fine-grained features, including color patterns, beak shape, wing structure, or other category-specific characteristics. 
%Such descriptions provide fine-grained semantic cues that explicitly correspond to object parts and attributes (e.g., beak shape, wing patterns). This makes language particularly useful not only for classification, but also for spatially grounding discriminative visual regions.
Moreover, language-based supervision aligns well with real-world recognition settings, where domain experts frequently rely on textual resources such as field guides, species descriptions, or taxonomic annotations to distinguish visually similar categories.

A common strategy for incorporating textual information in visual recognition is to use multimodal encoders~\cite{ruthardt2026steerable,radford2021learning}. However, these methods typically rely on textual inputs during inference, which is often infeasible in practical deployment settings. This raises the question: \textit{Can language still improve recognition if it is used only as a training-time signal that shapes visual representations?}

In addition to correctly distinguishing between visually similar categories, a fine-grained classifier should provide confidence estimates that meaningfully reflect its predictive reliability. This is especially relevant when several categories share similar visual characteristics: a model may assign high confidence to an incorrect fine-grained prediction, making it difficult to identify ambiguous or uncertain cases~\cite{venkataramanan2025probabilistic}. Classification accuracy alone does not capture this aspect of model behavior. Calibration provides a complementary perspective by measuring the agreement between predicted confidence and empirical prediction accuracy. Thus, improving fine-grained recognition also raises an important question: \textit{Can the supervision used to learn discriminative visual features improve the reliability of the resulting confidence estimates?}

Motivated by these questions, we study textual attributes under the Learning Using Privileged Information (LUPI) paradigm~\cite{vapnik2009new,vapnik2015learning}, where textual information is available only during training and is not used at inference time. Prior LUPI work uses privileged information to supervise attention~\cite{rodriguez2023fine} or improve detection~\cite{bartolo2026enhancing}. In this work, we formulate deep FGVC as a LUPI problem in which textual attribute descriptions provide training-time privileged semantic supervision. Unlike classical LUPI, where privileged information may provide additional information unavailable from the primary modality, our textual descriptions provide semantic guidance about the visual content during training. The text supervises a multimodal teacher toward fine-grained visual distinctions, whose knowledge is subsequently transferred to an image-only student through distillation. To the best of our knowledge, this is the first work to investigate such semantic textual supervision for deep FGVC under the LUPI setting.

We propose Fine-Grained Learning Using Privileged Information (FG-LUPI), a teacher-student framework built upon Vision Transformers (ViTs)~\cite{dosovitskiy2021an}. The teacher is extended into a multimodal model that learns from both images and textual descriptions, while the student operates only on visual inputs. By integrating textual attributes with visual representations via gated cross-attention, the teacher learns richer semantic representations that capture fine-grained discriminative features. This knowledge is then distilled into the image-only student, allowing it to inherit the teacher's language-guided attention patterns without requiring textual input at inference time.
We design a multi-level distillation framework that transfers the knowledge acquired by the multimodal teacher to the image-only student across multiple representation levels.  Using this framework, the student is trained to inherit: (i) global predictive behavior through prediction distillation, (ii) localized feature representations through patch-level distillation, and (iii) spatial relationships through attention distillation. Together, these objectives encourage the student to reproduce both the discriminative patch representations and the attention patterns learned by the multimodal teacher. %As a result, the student learns to localize and exploit subtle discriminative visual cues for fine-grained recognition, despite relying only on images at inference time. 
%A conceptual overview of the proposed framework is shown in Fig.~\ref{fig:lupi_illustration}. 

Experiments on multiple fine-grained recognition benchmarks, including CUB-200-2011 [59], NABirds [53], Stanford Dogs [25], FGVCAircraft [35], Stanford Cars [28], and iNaturalist 2021 [22], demonstrate that the proposed framework consistently improves performance over strong ViT-based baselines and competitive fine-grained classification methods. In addition to classification accuracy, we evaluate Expected Calibration Error (ECE) to examine the confidence calibration of the resulting image-only classifier. FG-LUPI achieves lower ECE than the vanilla ViT baseline across all six benchmarks, suggesting that the benefits of privileged textual supervision extend beyond predictive accuracy. Qualitative analyses of attention maps further show that the student focuses on meaningful and discriminative image regions (see Fig. 1), suggesting that text-guided attention patterns learned by the multimodal teacher are retained through distillation.

Our contributions are summarized as follows:
(1) We formulate fine-grained visual classification under the LUPI paradigm, treating textual attribute descriptions as privileged training-time information while preserving image-only inference.
(2) We propose FG-LUPI, a multimodal teacher-student framework that transfers knowledge acquired from textual attribute supervision to an image-only classifier through multi-level distillation.
(3) We demonstrate the effectiveness of textual attribute descriptions across diverse fine-grained recognition benchmarks, showing improvements in classification accuracy and consistently lower calibration error than vanilla ViT baselines.

\section{Related Works}

\textbf{Fine-grained Visual Classification.}
A prominent line of research in FGVC focuses on localizing discriminative object regions and learning part-level representations, either using part annotations~\cite{zhangPartBasedRCNNsFineGrained2014,huangPartStackedCNNFineGrained2016} or weakly supervised learning from image-level labels~\cite{geWeaklySupervisedComplementary2019,zhangWeaklySupervisedFineGrained2016,korsch2019classification,sun2024sim}. Beyond explicit part modeling, attention mechanisms have been widely adopted to identify discriminative image regions and capture their spatial relationships~\cite{fuLookCloserSee2017,pengObjectPartAttentionModel2018,sunMultiAttentionMultiClassConstraint2018}. More recently, ViTs~\cite{dosovitskiy2021an,touvron2021training} have become a dominant architecture for FGVC due to their ability to model long-range dependencies through self-attention. Building on this capability, several methods exploit transformer attention to select informative tokens and progressively refine discriminative representations across layers, leading to improved recognition performance~\cite{he2022transfg,wang2021feature,wang2025feature,shi2025ldh,yang2026revisiting,bi2025universal,wang2024multi}.

\textbf{Multimodal Methods for FGVC.}
Recent approaches incorporate auxiliary information such as textual descriptions. Vision-language models (VLMs)~\cite{yao2021filip,zhengDreamLIPLanguageImagePretraining2025,xiao2025flair} learn visual-semantic representations from large-scale image-text data, while other methods exploit textual or metadata information through cross-modal fusion~\cite{diaoMetaFormerUnifiedMeta2022,alayrac2022flamingo}. Alternate VLM-based approaches have also used language guidance to identify discriminative visual attributes~\cite{jiang2024delving,radford2021learning, xie2025fg, saha2024improved, he2017fine}. While these methods demonstrate the value of textual supervision, text generally remains part of the multimodal recognition process. FG-LUPI instead treats text as training-only privileged information and transfers the resulting language-guided knowledge to an image-only model.

\textbf{Knowledge Distillation.}
Knowledge distillation (KD) transfers knowledge from a teacher to a student through output predictions~\cite{hinton2015distilling} or intermediate representations~\cite{romero6550fitnets}. Subsequent approaches have explored attention~\cite{zagoruyko2016paying}, relational~\cite{tung2019similarity}, contrastive~\cite{tian2019contrastive}, and decoupled logit-level distillation~\cite{zhao2022decoupled}. KD has also been applied to ViTs~\cite{touvron2021training} and FGVC, where teacher supervision is used to transfer discriminative part information and intermediate representations~\cite{zheng2019looking,liu2020filtration,zhang2025dual}. These approaches, however, typically distill knowledge between models operating on the same modality. FG-LUPI instead distills knowledge from a teacher conditioned on privileged textual attributes into an image-only student, combining prediction-, patch-, and attention-level supervision.

\begin{figure*}[t]
    \centering
    \includegraphics[width=0.9\linewidth]{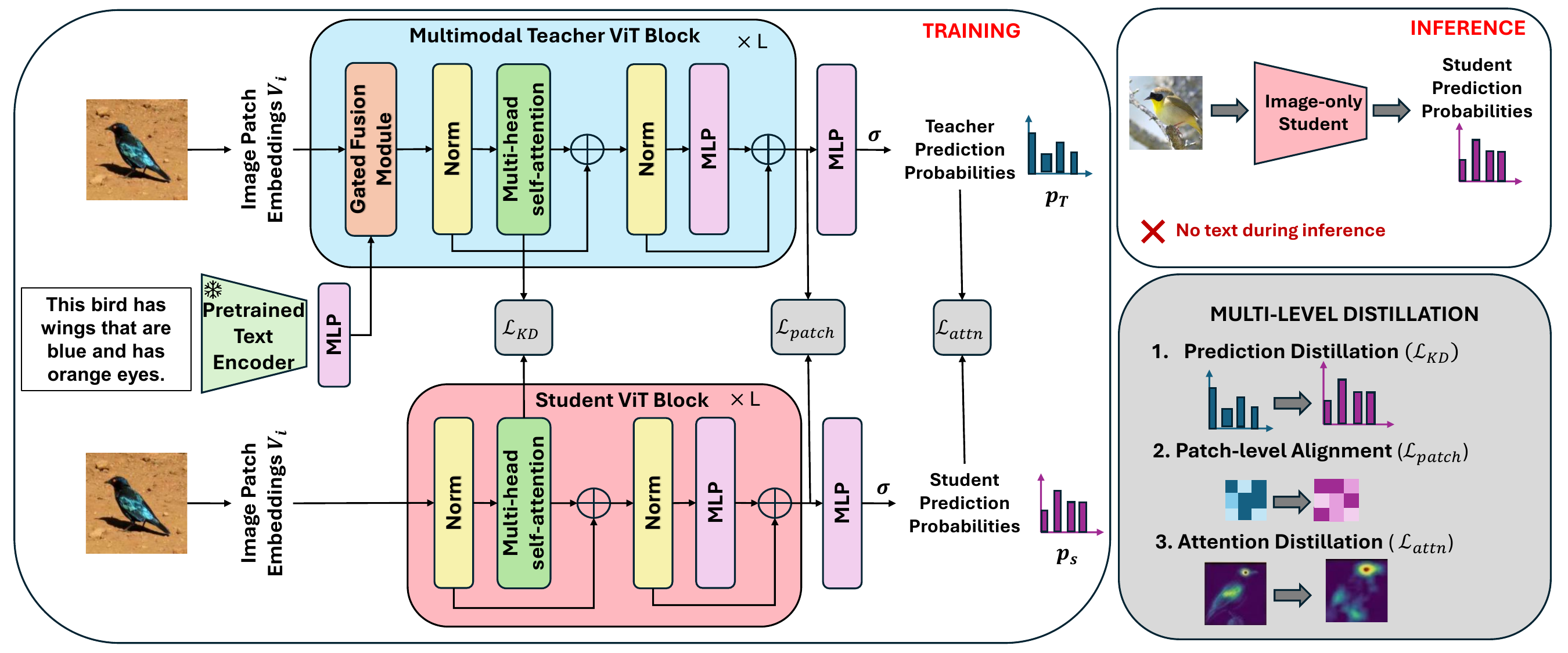}
\caption{\textbf{Architecture of the proposed FG-LUPI framework.} During training, a multimodal teacher receives both images and textual attribute descriptions. Textual features are integrated into the visual stream through gated cross-attention layers, enabling the teacher to learn joint visual-semantic representations under multimodal supervision. An image-only student with the same ViT backbone is trained using prediction, patch-level, and attention distillation objectives. These objectives transfer predictive behavior, intermediate feature representations, and attention structure from the teacher to the student. During inference, only the student model is retained.}

    \label{fig:architecture}
\end{figure*}

\textbf{Learning Using Privileged Information.}
Learning Using Privileged Information (LUPI) exploits auxiliary information available during training but unavailable at inference~\cite{vapnik2009new,vapnik2015learning}. It has been applied to image retrieval~\cite{sharmanska2013learning,chen2020learning}, classification~\cite{fu2023coarse,niu2018fine}, detection~\cite{yang2017miml,bartolo2026enhancing}, and segmentation~\cite{lee2018spigan}. In fine-grained recognition, privileged keypoint annotations have been used to supervise attention~\cite{rodriguez2023fine}, while other LUPI approaches combine privileged modalities with knowledge distillation~\cite{lopez2015unifying,hajavi2023audio,bartolo2026enhancing}. FG-LUPI extends this to textual attribute descriptions to guide a multimodal teacher and distilling the resulting language-guided visual knowledge into an image-only student.

Existing multimodal approaches exploit text directly, while conventional distillation generally transfers knowledge from visually trained teachers. FG-LUPI combines these directions by using text as training-only privileged information and transferring its semantic guidance to an image-only model through multi-level distillation.

\section{Method}

\subsection{Problem Setting}

Let $\mathcal{D}_{\text{train}}={(x_i,s_i,y_i)}_{i=1}^{n}$ denote the training set, where $x_i \in \mathbb{R}^{H \times W \times 3}$ is an input image, $s_i$ is a corresponding textual attribute description, and $y_i \in \{1,\dots,C\}$ is the class label among $C$ fine-grained categories. We formulate fine-grained classification within the LUPI paradigm, where textual descriptions are treated as privileged information available only during training.

As illustrated in Fig.~\ref{fig:architecture}, FG-LUPI consists of a multimodal teacher model $f_T(\cdot,\cdot)$ and an image-only student model $f_S(\cdot)$. The teacher receives both image and text inputs and learns joint visual-semantic representations from textual attribute descriptions and image content. In contrast, the student operates solely on images and is designed to perform classification without access to textual information.
The knowledge learned by the teacher is transferred to the student through distillation objectives that encourage alignment at multiple representation levels, including class predictions, patch-level representations, and self-attention patterns. Through this process, the student learns to benefit from the semantic information available during training while maintaining image-only inference.

\subsection{Multimodal Teacher Network}

The teacher network incorporates textual attribute information into intermediate visual representations during training through gated cross-attention. The model is built upon a ViT backbone with $L$ layers and augmented with a textual branch that provides semantic conditioning through gated cross-attention layers.

Given an input image $x_i$, the vision encoder produces a sequence of patch embeddings $V_i = \{v_{i, \text{cls}}, v_{i, 1}, v_{i, 2}, \dots, v_{i, P}\} \in \mathbb{R}^{(P+1) \times d_v},$
where $v_{i, \text{cls}}$ denotes the class token, $P$ is the number of patches and $d_v$ is the visual embedding dimension. In parallel, a textual description $s_i$ is encoded using a frozen pretrained language model, yielding contextual token embeddings $T_i = \{t_{i,1}, t_{i,2}, \dots, t_{i,M}\} \in \mathbb{R}^{M \times d_t}$, where $M$ denotes the number of text tokens and $d_t$ is the dimension of the text token embedding. The textual representations are subsequently projected into the image embedding space using a learnable projection layer $\phi(\cdot)$, yielding $\bar{T_i} \in \mathbb{R}^{M \times d_v}$.

Let $V_i^{(l)}$ denote the visual token representations at transformer layer $l$. To inject textual information into the visual stream, we employ standard multi-head cross-attention, with the visual tokens
serving as queries and the textual tokens serving as keys and values:
\begin{equation}
\mathrm{MHA}(V_i,\bar{T}_i,\bar{T}_i)
=
\mathrm{Concat}(\mathrm{head}_1,\ldots,\mathrm{head}_h)W^O,
\label{eq:cross_attention}
\end{equation}
where
\begin{equation}
\mathrm{head}_j =
\sigma\left(
\frac{(V_iW_j^Q)(\bar{T}_iW_j^K)^\top}
{\sqrt{d_k}}
\right)
(\bar{T}_iW_j^V).
\end{equation}
Here, $\sigma(\cdot)$ is the softmax function, $W_j^Q$, $W_j^K$, and $W_j^V$ are the learnable query, key, and
value projections for attention head $j$, respectively, $W^O$ is the
output projection, $h$ is the number of attention heads, and $d_k$ is
the key dimension.

Following recent multimodal fusion strategies~\cite{ruthardt2026steerable, alayrac2022flamingo}, this interaction is integrated into the visual backbone through a gated residual connection:
\begin{equation}
\tilde{V}_i^{(l)}
=
V_i^{(l)}
+
\alpha^{(l)}
\mathrm{MHA}\left(
V_i^{(l)},\bar{T}_i,\bar{T}_i
\right),
\label{eq:gated_fusion}
\end{equation}
where $\alpha^{(l)}$ is a learnable scalar controlling the contribution
of the textual information at layer $l$.

The teacher is trained using the standard cross-entropy loss:
\begin{equation}\label{eq:teacher_loss}
\mathcal{L}_{\text{T}}
=
-\log p_{T,y_i},
\end{equation}
where $p_T = \sigma\!\left(f_T(x_i,s_i)\right)$
is the teacher's predicted class probability distribution.

\subsection{Image-Only Student Network}

The student network is designed to learn from the multimodal teacher while operating only on images at inference time. Architecturally, it adopts the same ViT backbone as the teacher but removes the textual encoder and cross-attention modules.
To transfer the teacher's knowledge, the student is optimized using multi-level distillation objectives. These objectives encourage the student to align its predictions, patch-level representations, and attention patterns with those of the teacher. %The specific distillation objectives are described in the following subsections.

\paragraph{Prediction Distillation.}
To transfer the teacher’s predictive behavior, we employ prediction distillation. Let $z_T = f_T(x_i, s_i)$ and $z_S = f_S(x_i)$ denote the logits of the teacher and student, respectively. We define the softened probability distributions as:
\begin{equation}
p_T^\tau = \sigma\!\left(\frac{z_T}{\tau}\right), \quad
p_S^\tau = \sigma\!\left(\frac{z_S}{\tau}\right),
\end{equation}
where $\tau$ is a temperature parameter controlling the smoothness of the distributions. % and $\sigma(\cdot)$ is the softmax function.
The predictive distillation loss is defined as:
\begin{equation}
\mathcal{L}_{\mathrm{KD}} =
\mathrm{KL}\!\left(p_T^\tau \,\|\, p_S^\tau\right).
\end{equation}

\paragraph{Patch-Level Distillation.}
While prediction distillation transfers global predictive behavior, it does not explicitly constrain the spatial structure of intermediate representations. To address this, we introduce a patch-level distillation objective that aligns patch embeddings between the teacher and student.

Let $P_S, P_T \in \mathbb{R}^{N \times D}$ denote the final patch embeddings from the last transformer block of the student and teacher, respectively, where $N$ is the number of patches and $D$ is the embedding dimension. Since both models share the same ViT patchification scheme, a one-to-one correspondence between spatial tokens is possible. We denote by $P_{S,n}$ and $P_{T,n}$ the $n$-th row (patch token) of $P_S$ and $P_T$.

We apply a patch-wise cosine alignment loss:
\begin{equation}
\mathcal{L}_{\text{patch}}
=
\frac{1}{N}
\sum_{n=1}^{N}
\left(
1 -
\frac{
P_{S,n}^\top P_{T,n}
}{
\|P_{S,n}\|_2 \, \|P_{T,n}\|_2
}
\right),
\end{equation}
which encourages the student to align its patch-level representations with those of the teacher. Cosine similarity is adopted here to enforce scale-invariant alignment of patch embeddings, rather than absolute feature values.

\paragraph{Attention Distillation.}
In addition to feature-level alignment, we transfer the teacher’s spatial interaction patterns through attention distillation. Since the teacher is conditioned on textual attributes, its attention maps reflect interactions learned under multimodal supervision.

Let
$A_T^{(l)}, A_S^{(l)} \in \mathbb{R}^{h \times N \times N}$
denote the patch-to-patch attention maps at layer $l$, obtained after removing the class token, where $h$ is the number of attention heads and $N$ is the number of patch tokens. The attention weights remain valid probability distributions, as each row is produced via a softmax over patch tokens. We denote by $\hat{A}_T^{(l)}$ and $\hat{A}_S^{(l)}$ the corresponding row-wise normalized attention maps. The attention distillation loss is defined as:
\begin{equation}
\mathcal{L}_{\text{attn}}
=
\frac{1}{L}
\sum_{l=1}^{L}
\frac{1}{h}
\sum_{k=1}^{h}
\mathrm{KL}
\left(
\hat{A}_{T,k}^{(l)} \;\|\; \hat{A}_{S,k}^{(l)}
\right),
\end{equation}
where KL divergence is computed row-wise over patch tokens.
This objective encourages the student to reproduce the attention patterns learned by the teacher, thereby transferring relational information captured during multimodal training.

\paragraph{Student Training Objective.}
In addition to distillation, the student is supervised using ground-truth labels via a standard cross-entropy loss:
\begin{equation}\label{eq:student_loss}
\mathcal{L}_{\mathrm{CE}}^{S} = -\log p_{S,y_i},
\end{equation}
where $p_S = \sigma(z_S)$ and $z_S = f_S(x_i)$ denotes the student logit vector.
The final student objective combines cross-entropy supervision with all distillation components:
\begin{equation}
\mathcal{L}_{S}
=
\mathcal{L}_{\mathrm{CE}}^{S}
+
\lambda_1 \mathcal{L}_{\mathrm{KD}}
+
\lambda_2 \mathcal{L}_{\text{patch}}
+
\lambda_3 \mathcal{L}_{\text{attn}},
\end{equation}
where $\lambda_1$, $\lambda_2$, and $\lambda_3$ are weighting parameters for balancing the three distillation objectives.

\subsection{Training and Inference}

The teacher and the student are trained simultaneously in an online distillation setting. The teacher is optimized using the classification loss in Eq.~\ref{eq:teacher_loss}, while the student is optimized using the overall objective in Eq.~\ref{eq:student_loss}. Because both networks are updated throughout training, the teacher's predictions and intermediate representations form non-stationary distillation targets.
To stabilize these targets, we maintain an exponential moving average (EMA)~\cite{tarvainen2017mean} of the teacher parameters. Let $\theta_T^{t}$ denote the teacher parameters at iteration $t$, and $\theta_{\text{EMA}}^{t}$ denote the EMA parameters, which are updated as:
\begin{equation}
\theta_{\text{EMA}}^{t}
=
\mu \theta_{\text{EMA}}^{t-1}
+
(1-\mu)\theta_T^{t},
\end{equation}
where $\mu \in [0,1)$ is the momentum coefficient.
The EMA teacher provides the distillation targets for prediction, patch-level, and attention distillation. During inference, only the student network is retained.

\section{Experiments}

\subsection{Experimental Setup}\label{sec:exp}

\textbf{Datasets.}
We evaluate our approach on CUB-200-2011~\cite{wah2011caltech}, NABirds~\cite{van2015building}, Stanford Dogs~\cite{khosla2011novel}, FGVCAircraft~\cite{maji2013fine}, Stanford Cars~\cite{krause2013collecting}, and iNaturalist 2021 (mini split) ~\cite{inaturalist21}. 
For CUB-200-2011, we use the human-authored descriptions~\cite{reedLearningDeepRepresentations2016} as privileged information. For the remaining datasets, descriptions are generated using Gemma 4 E4B~\cite{google2026gemma4}. These descriptions serve as training-time privileged semantic supervision, providing a complementary semantic representation of the visual observations that exposes potentially discriminative attributes to the multimodal teacher. This semantic supervision is subsequently transferred to the image-only student through distillation.
Full prompts are provided in Appendix~\ref{app:dataset}.

\begin{table}[t]
    \begin{adjustbox}{width=0.95\linewidth, center}
    \centering
    \begin{tabular}{ccccc} \toprule
      Method & Backbone & CUB & Dogs  & NABirds \\ \midrule
      Cross-X~\cite{luo2019cross} & ResNet-50 & 88.7 & 88.9 & 86.2 \\
      PMG~\cite{zhang2022learning} & ResNet-101 & 90.0 & 90.7 & 88.4 \\
      SDNs~\cite{zhuang2020learning} & ResNet-101 & 88.8 & - & 88.3 \\
      API-Net~\cite{xu2023fine} & DenseNet-161 & 90.0 & 89.4 & 88.1 \\
      %DPD-Net~\cite{zhang2025dual} & ResNet-50 & 88.1 & - & - \\
      \midrule
      ViT~\cite{dosovitskiy2021an}   & ViT-B/16 & 90.3 & 91.6 &  89.6\\
      DeiT~\cite{touvron2021training} & DeiT-B/16 & 87.6 & 91.5 & 87.1 \\
      TransFG~\cite{he2022transfg}   & ViT-B/16 & 91.7 & 92.3 &  90.8\\
      IELT~\cite{xu2023fine} & ViT-B/16 & 91.8 & 91.8 & 90.8 \\
      MP-FGVC~\cite{jiang2024delving} & ViT-B/16 & 91.8 & 91.0 & 91.0 \\
      ACC-ViT~\cite{zhang2024vision} & ViT-B/16 & 91.4 & 92.1 & 90.9 \\
      FEMCC~\cite{wang2021feature} & ViT-B/16 & 89.3 & - & 89.4 \\
      LDH-ViT~\cite{shi2025ldh} & ViT-B/16 & 91.7 & 93.1 & 90.2 \\
      $A^2$Q-Trans~\cite{qiao2025fine} & ViT-B/16 & \underline{92.4} & \underline{94.9} & \underline{91.3} \\ \midrule
      FG-LUPI Teacher & ViT-B/16 & 91.4$\pm$0.1 & 93.0$\pm$0.2 & 91.8$\pm$0.1 \\
      FG-LUPI Student & ViT-B/16 & \textbf{92.8$\pm$0.2} & \textbf{95.1$\pm$0.1} & \textbf{92.3$\pm$0.2}\\ \bottomrule
    \end{tabular}
\end{adjustbox}
    \caption{\textbf{Top-1 classification accuracy (\%) on CUB-200-2011, Stanford Dogs, and NABirds.} FG-LUPI consistently demonstrates superior performance over CNN- and ViT-based methods. The best scores are in bold and the
    second-best are underlined.}
    \label{tab:results_1}
\end{table}

\begin{figure*}[ht]
\centering
\setlength{\tabcolsep}{0pt}
\renewcommand{\arraystretch}{0.95}

\begin{tabular}{@{}c c c @{\hspace{4pt}} c c @{\hspace{4pt}} c c@{}}

\rotatebox{90}{\makebox[2.2cm][c]{\small NABirds}} &
\includegraphics[width=2.3cm,height=2.3cm]{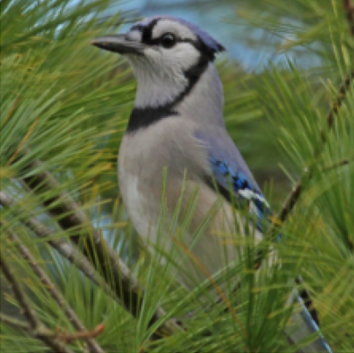} & 
\includegraphics[width=2.3cm,height=2.3cm]{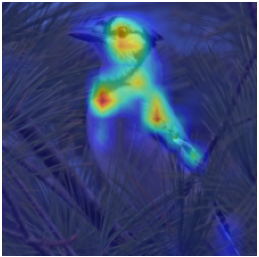} &
\includegraphics[width=2.35cm,height=2.3cm]{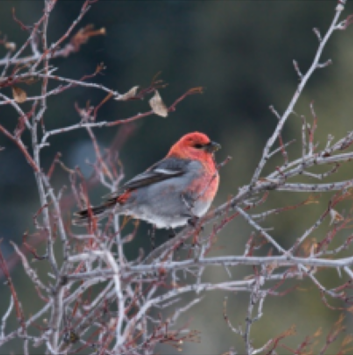} &
\includegraphics[width=2.3cm,height=2.3cm]{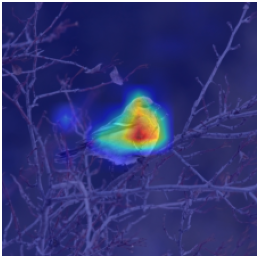} &
\includegraphics[width=2.3cm,height=2.3cm]{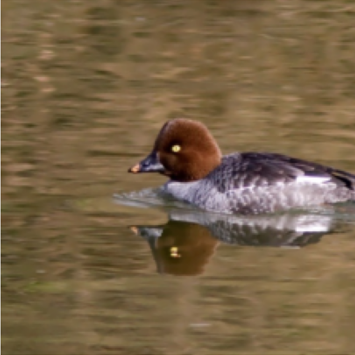} & 
\includegraphics[width=2.3cm,height=2.3cm]{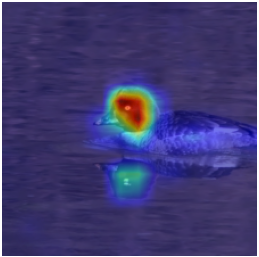}\\

\rotatebox{90}{\makebox[2.2cm][c]{\small Cars}} &
\includegraphics[width=2.3cm,height=2.3cm]{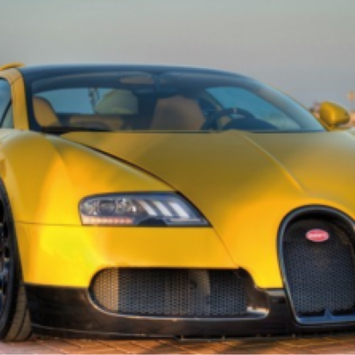} & 
\includegraphics[width=2.3cm,height=2.3cm]{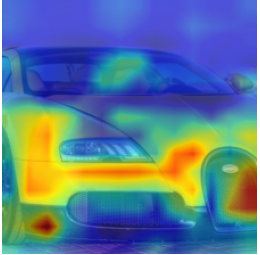} &
\includegraphics[width=2.3cm,height=2.3cm]{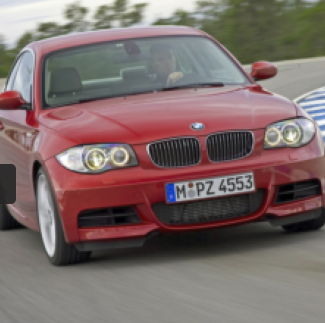} &
\includegraphics[width=2.3cm,height=2.3cm]{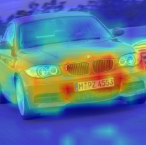} &
\includegraphics[width=2.3cm,height=2.3cm]{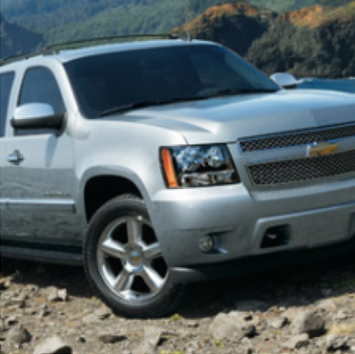} & 
\includegraphics[width=2.3cm,height=2.3cm]{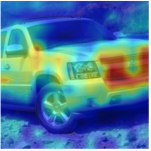}\\

\rotatebox{90}{\makebox[2.2cm][c]{\small Aircraft}} &
\includegraphics[width=2.3cm,height=2.3cm]{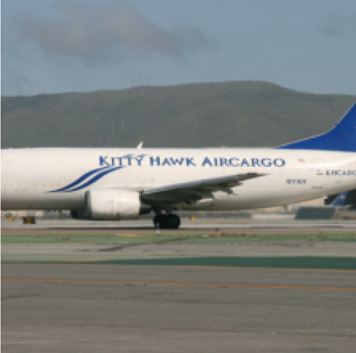} &
\includegraphics[width=2.3cm,height=2.3cm]{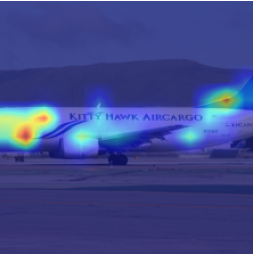} &
\includegraphics[width=2.35cm,height=2.3cm]{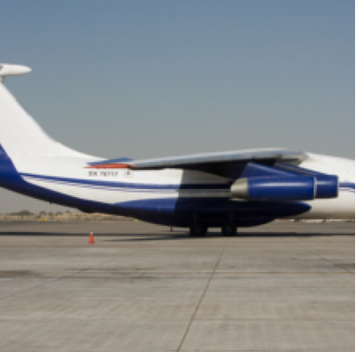} &
\includegraphics[width=2.3cm,height=2.3cm]{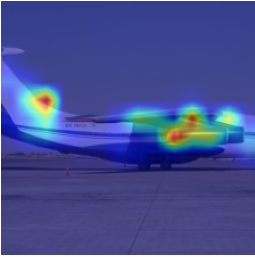} &
\includegraphics[width=2.3cm,height=2.3cm]{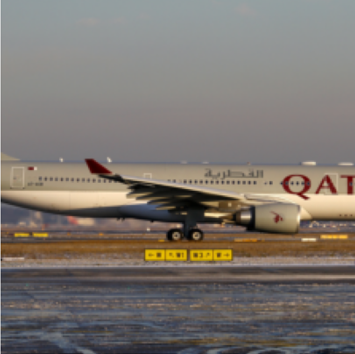} & 
\includegraphics[width=2.3cm,height=2.3cm]{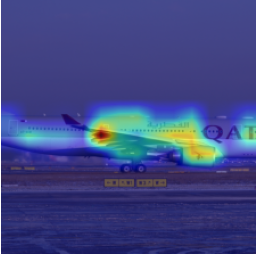}\\

 \rotatebox{90}{\makebox[2.2cm][c]{\small Dogs}} &
\includegraphics[width=2.3cm,height=2.3cm]{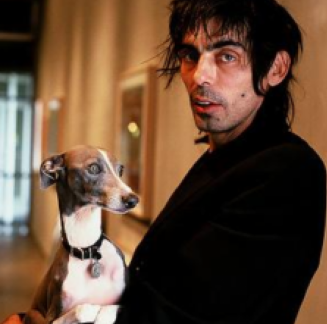} & 
\includegraphics[width=2.3cm,height=2.3cm]{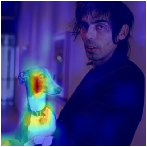} &
\includegraphics[width=2.35cm,height=2.3cm]{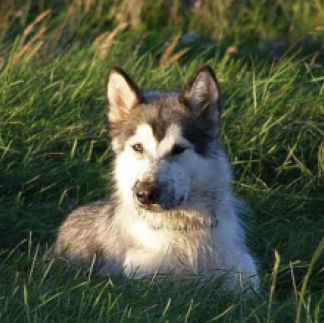} &
\includegraphics[width=2.3cm,height=2.3cm]{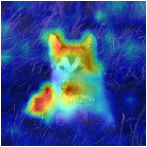} &
\includegraphics[width=2.3cm,height=2.3cm]{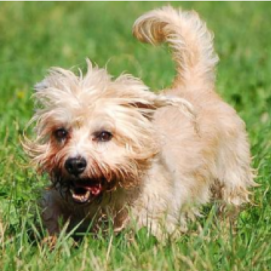} & 
\includegraphics[width=2.3cm,height=2.3cm]{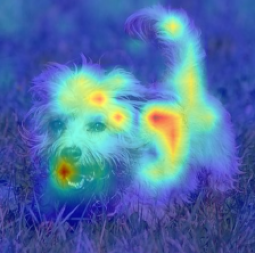}\\

\end{tabular}

\caption{\textbf{Visualization of attention maps generated by FG-LUPI across multiple fine-grained recognition datasets.} For each example, the left image shows the original input and the right image shows the corresponding attention map. The attention maps generally concentrate on visually relevant object regions while reducing attention to surrounding background regions.}
\label{fig:attn_viz}
\end{figure*}

\textbf{Implementation Details.}
Both the teacher and student are based on the ViT-B/16 architecture initialized with ImageNet-pretrained weights~\cite{deng2009imagenet}. For the multimodal teacher, textual descriptions are encoded using a pretrained BERT encoder~\cite{devlin2019bert} with frozen parameters. Following baseline approaches, input images are resized to $448 \times 448$ pixels for all datasets except iNaturalist 2021, where images are resized to $304 \times 304$. Image augmentations during training include random resized cropping and horizontal flipping, while center cropping is used during evaluation. For comparisons with prior work, we report the performance numbers from the corresponding publications.

Optimization is performed using AdamW for a total of $50$ epochs with a batch size of $16$. Separate optimizers are maintained for the teacher and student. For the teacher, the vision backbone uses a learning rate of $10^{-6}$, while the projection layers, cross-attention fusion blocks, and final classification head are optimized with a learning rate of $10^{-4}$. The student network has a learning rate of $10^{-5}$ for the backbone and $10^{-4}$ for the student projection and classification heads. The EMA teacher is updated using a decay rate of $\mu = 0.999$. The student distillation objective has equal weighting parameters  of $\lambda_1 = 0.5$, $\lambda_2 = 0.5$, $\lambda_3 = 0.5$.

\subsection{Results}

\textbf{Quantitative Results.} Table~\ref{tab:results_1} compares FG-LUPI with CNN- and ViT-based fine-grained classification methods on CUB-200-2011, Stanford Dogs, and NABirds. FGVC benchmarks such as these are well-established, and recent methods often yield incremental improvements. FG-LUPI consistently achieves state-of-the-art accuracy across all three datasets. Compared with the vanilla ViT-B/16 baseline, FG-LUPI improves performance by 2.5 percentage points on CUB, 3.5 percentage points on Stanford Dogs, and 2.7 percentage points on NABirds, demonstrating consistent gains from incorporating privileged textual information during training.
Results on FGVCAircraft and Stanford Cars are reported in Table~\ref{tab:results_2}. FG-LUPI achieves the highest accuracy on FGVCAircraft. On Stanford Cars, FG-LUPI achieves the second-highest accuracy of 89.1\%, while substantially improving over the vanilla ViT-B/16 baseline by 5.5 percentage points. The remaining gap to the strongest baseline method may reflect the limitations of the available textual descriptions, which may emphasize broader semantic properties rather than the highly localized and model-specific cues required to distinguish visually similar car models.
Table~\ref{tab:inat21} presents results on the iNaturalist 2021 benchmark. FG-LUPI improves top-1 accuracy from 74.08\% to 76.46\% over the vanilla ViT-B/16 baseline. While the gain is smaller than on several other benchmarks, the privileged information available for iNaturalist consists of class-level textual descriptions rather than instance-level attribute annotations. Consequently, the multimodal teacher receives less image-specific semantic supervision, which may limit the amount of fine-grained knowledge that can be transferred to the image-only student.

\begin{table}[]

    \begin{adjustbox}{width=0.75\linewidth, center}
    \centering
    \begin{tabular}{cccc} \toprule
      Method & Backbone & Aircraft & Cars \\ \midrule
      ViT~\cite{dosovitskiy2021an}   & ViT-B/16 & 90.3 & 83.6 \\
      TransFG~\cite{he2022transfg}   & ViT-B/16 & 90.9 & 84.6\\
      ACC-ViT~\cite{zhang2024vision} & ViT-B/16 & 91.8 & 84.1 \\
      $A^2$Q-Trans~\cite{qiao2025fine} & ViT-B/16 & \underline{93.6} & \textbf{90.8} \\ \midrule
      FG-LUPI Teacher & ViT-B/16 & 92.6$\pm$0.2 & 87.0$\pm$0.2 \\
      FG-LUPI Student & ViT-B/16 & \textbf{94.0$\pm$0.2} & \underline{89.2$\pm$0.1}\\ \bottomrule
    \end{tabular}
\end{adjustbox}
    \caption{\textbf{Top-1 classification accuracy (\%) on FGVCAircraft and Stanford Cars.} FG-LUPI achieves the highest accuracy on FGVCAircraft while maintaining strong performance on Stanford Cars. The best scores are in bold and the
    second-best are underlined.}
    \label{tab:results_2}
\end{table}

\begin{table}[]

\begin{adjustbox}{width=0.8\linewidth, center}
    \centering
    \begin{tabular}{cccc} \toprule
     Backbone   & Vanilla & FG-LUPI Teacher & FG-LUPI Student\\ \midrule
     ViT-B/16 & 74.08 & 76.34 & \textbf{76.46} \\ \bottomrule   
    \end{tabular}
\end{adjustbox}
\caption{\textbf{Top-1 classification accuracy (\%) on iNaturalist 2021.} Comparison between the vanilla ViT-B/16 baseline and FG-LUPI.}
\label{tab:inat21}
\end{table}

\begin{table}[]

\begin{adjustbox}{width=0.95\linewidth, center}
    \centering
    \begin{tabular}{cccccc} \toprule
     Model   & CUB & Dogs & NABirds & Aircraft & Cars  \\ \midrule
     FitNet~\cite{romero6550fitnets} & 90.2$\pm$0.1 & 93.4$\pm$0.2 & 90.0$\pm$0.1 & 91.4$\pm$0.1 & 86.6$\pm$0.2    \\ 
     Attn Trans~\cite{zagoruyko2016paying} & 89.7$\pm$0.2 & 93.0$\pm$0.1 & 89.8$\pm$0.2 & 92.4$\pm$0.1 & 86.4$\pm$0.1   \\ 
     SP~\cite{tung2019similarity} & 90.9$\pm$0.1 & 93.1$\pm$0.1 & 90.1$\pm$0.2 & 92.2$\pm$0.1 & 86.8$\pm$0.2    \\
     DKD~\cite{zhao2022decoupled} & 91.5$\pm$0.1 & 93.6$\pm$0.2  & 90.2$\pm$0.1 & 93.0$\pm$0.1 &  87.2$\pm$0.2   \\
     CRD~\cite{tian2019contrastive} & 90.6$\pm$0.1 & 92.9$\pm$0.2 & 90.2$\pm$0.2 & 91.7$\pm$0.1 &  86.8$\pm$0.1   \\
     FG-LUPI & \textbf{92.8$\pm$0.2} & \textbf{95.1$\pm$0.1} & \textbf{92.3$\pm$0.2} & \textbf{94.0$\pm$0.2} & \textbf{89.2$\pm$0.1}  \\ \bottomrule   
    \end{tabular}
\end{adjustbox}
\caption{\textbf{Comparison with knowledge distillation methods.} All methods use the same student and teacher backbone and training setting for a fair comparison. The best performance in each dataset is shown in bold.}
\label{tab:distillation}
\end{table}
 %, corresponding to a gain of 2.38\%.

Interestingly, the FG-LUPI student consistently outperforms its multimodal teacher across all evaluated datasets. One explanation is that the student benefits from its own visual supervision while additionally receiving complementary semantic guidance from the teacher through prediction, patch, and attention-level distillation. This language-guided supervision can regularize the learning of more discriminative visual representations, allowing the image-only student to surpass the teacher while requiring no textual information at inference.

To further assess the effectiveness of our proposed distillation framework, we compare FG-LUPI with representative knowledge distillation methods, including FitNet~\cite{romero6550fitnets}, Attention Transfer (Attn Trans)~\cite{zagoruyko2016paying}, Similarity-Preserving (SP) distillation~\cite{tung2019similarity}, Decoupled Knowledge Distillation (DKD)~\cite{zhao2022decoupled}, and Contrastive Representation Distillation (CRD)~\cite{tian2019contrastive}. For a controlled comparison, all distillation baselines use the same teacher and student architectures as FG-LUPI, with the corresponding distillation objective being the only component that is changed. As shown in Table~\ref{tab:distillation}, FG-LUPI  outperforms the baseline distillation approaches. 
These results suggest that the performance gains are not simply attributable to applying knowledge distillation to the multimodal teacher, but to the proposed multi-level distillation strategy, which jointly transfers complementary information at the prediction, patch, and attention levels to better preserve fine-grained visual knowledge in the image-only student.

\begin{table}[]

\begin{adjustbox}{width=0.95\linewidth, center}
    \centering
    \begin{tabular}{ccccccc} \toprule
     Model   & CUB & Dogs & NABirds & Aircraft & Cars & iNat21 \\ \midrule
     ViT & 5.44 & 5.69 & 15.35 & 7.64 & 8.74  & 22.32 \\ 
     FG-LUPI & \textbf{1.92} & \textbf{2.54} & \textbf{9.06} & \textbf{4.30} & \textbf{4.61} & \textbf{11.29} \\ \bottomrule   
    \end{tabular}
\end{adjustbox}
\caption{\textbf{Expected Calibration Error (ECE, \%) across fine-grained classification benchmarks.} Lower values indicate better calibration. FG-LUPI consistently achieves substantially lower calibration error than the vanilla ViT baseline across all datasets.}
\label{tab:ece}
\end{table}

%In addition to improving classification accuracy, FG-LUPI also yields substantially better calibrated predictions. 
Table~\ref{tab:ece} reports the Expected Calibration Error (ECE) across all evaluated datasets. ECE quantifies the miscalibration of a model by measuring the average absolute difference between its predicted confidence scores and its actual classification accuracy. Therefore, a lower ECE indicates a better performance.
Compared to the vanilla ViT baseline, FG-LUPI consistently reduces calibration error. These results suggest that attribute-guided distillation not only improves classification performance but also produces more reliable confidence estimates. 

Additional comparison with VLMs (CLIP~\cite{radford2021learning}, SigLIP~\cite{zhai2023sigmoid}, CoCa~\cite{yu2022coca}, VLM2Vec~\cite{jiang2024vlm2vec} and BLIP~\cite{li2022blip}), and a discussion on computational expense is provided in Appendix~\ref{app:additional_results}.

\textbf{Qualitative Results.}
Fig.~\ref{fig:vit_vs_ours} and Fig.~\ref{fig:attn_viz} visualize the attention maps produced by FG-LUPI across different datasets. Overall, FG-LUPI attends to semantically meaningful regions such as object parts and structural details, while reducing attention on background clutter.
On the CUB-200-2011 dataset (Fig.~\ref{fig:vit_vs_ours}), the vanilla ViT tends to focus on both the bird and surrounding background regions, whereas FG-LUPI shifts attention more precisely to the bird regions. On NABirds (Fig.~\ref{fig:attn_viz}), the model suppresses irrelevant contextual elements such as branches and water, focusing instead on the bird itself. Similarly, on Stanford Dogs, when the image contains multiple objects or a human-dog interaction, FG-LUPI primarily attends to the dog region.
These visualizations suggest that the proposed distillation framework encourages the student to learn more discriminative features that are aligned with fine-grained semantic cues, despite operating without textual input at inference time. Additional qualitative analyses with vanilla ViT is provided in Appendix~\ref{app:additional_results}.

\begin{figure}[t]
\begin{tabular}{@{}ccc@{}} 
Original  & Teacher & Student \\
\includegraphics[width=0.3\linewidth]{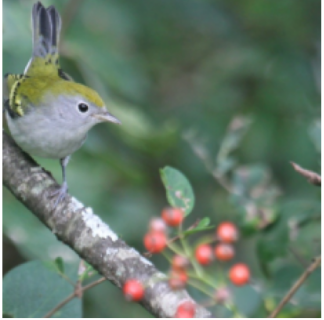} &
\includegraphics[width=0.3\linewidth]{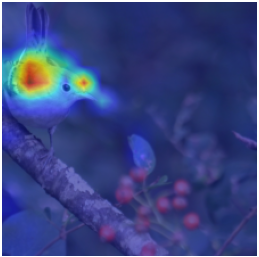} &
\includegraphics[width=0.3\linewidth]{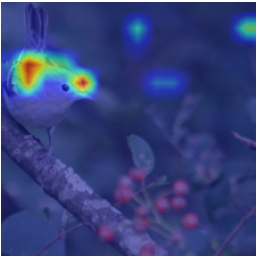}   \\
\includegraphics[width=0.3\linewidth]{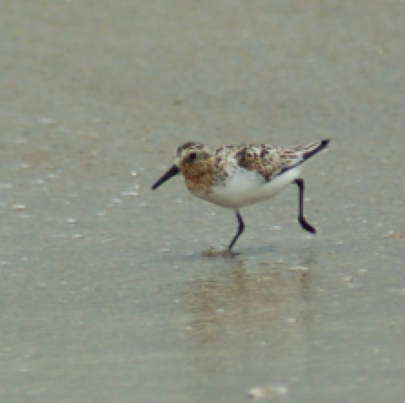} &
\includegraphics[width=0.3\linewidth]{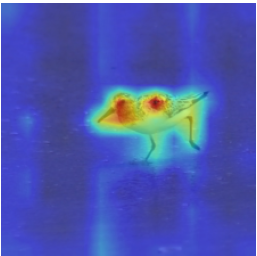} &
\includegraphics[width=0.3\linewidth]{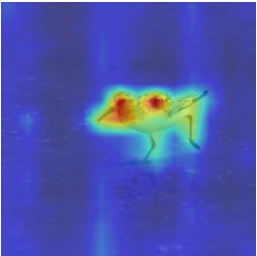}\\
\includegraphics[width=0.3\linewidth]{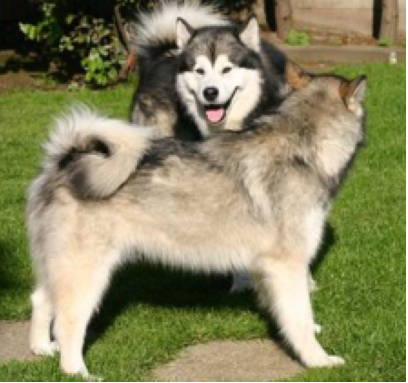} &
\includegraphics[width=0.3\linewidth]{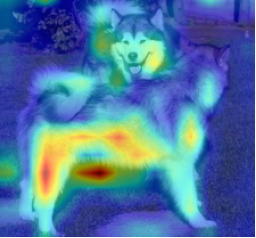} &
\includegraphics[width=0.3\linewidth]{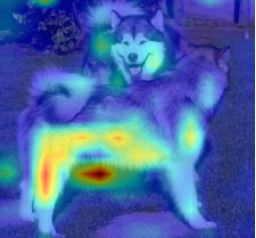}\\

\end{tabular}

\caption{\textbf{Visualization of attention maps from the multimodal teacher and the distilled image-only student.} Despite receiving only image inputs, the student exhibits attention patterns that closely resemble those of the teacher. This qualitative result suggests that the proposed distillation objectives effectively transfer information captured by the multimodal teacher.}
\label{fig:teacher_vs_student}
\end{figure}

We further analyze the effect of our distillation framework by comparing attention maps from the multimodal teacher and the image-only student in Fig.~\ref{fig:teacher_vs_student}. The teacher produces highly structured attention maps that are strongly concentrated on semantically relevant bird regions. We observe that the student is able to closely replicate these spatial focus patterns despite operating without textual inputs at inference time. This suggests that the proposed combination of prediction, patch-level, and attention distillation effectively transfers the teacher’s attribute-guided localization behavior into the student model.
 %This behavior is consistent with token-level cross-attention mechanisms and the inherently non-disjoint structure of fine-grained visual attributes. 
Additional qualitative results on cross-dataset attention generalization and text-conditioned attention is provided in Appendix~\ref{app:additional_results}.

\subsection{Ablation Study}

We conduct all ablation experiments on CUB-200-2011, using the same experimental protocol as described in Sec.~\ref{sec:exp}.

\textbf{Effect of Distillation Objectives.}
Table~\ref{tab:distillation_losses} evaluates the contribution of the proposed distillation losses. Among the individual objectives, prediction distillation ($\mathcal{L}_{KD}$) provides the largest improvement, achieving 91.64\% accuracy, while patch-level and attention distillation alone obtain 90.82\% and 90.46\%, respectively. Combining prediction distillation with either patch-level or attention supervision further improves performance to 92.42\% and 92.15\%, indicating that intermediate representation alignment provides complementary information beyond output-level matching. In contrast, combining only patch-level and attention distillation without prediction supervision yields a smaller gain (91.75\%), highlighting the importance of transferring the teacher's predictive behavior. Finally, integrating all three distillation objectives achieves the best performance of 93.03\%, suggesting that prediction, feature, and attention alignment capture complementary aspects of the teacher's knowledge and jointly contribute to effective knowledge transfer.

\begin{figure}[t]
    \centering
    \includegraphics[width=0.95\linewidth]{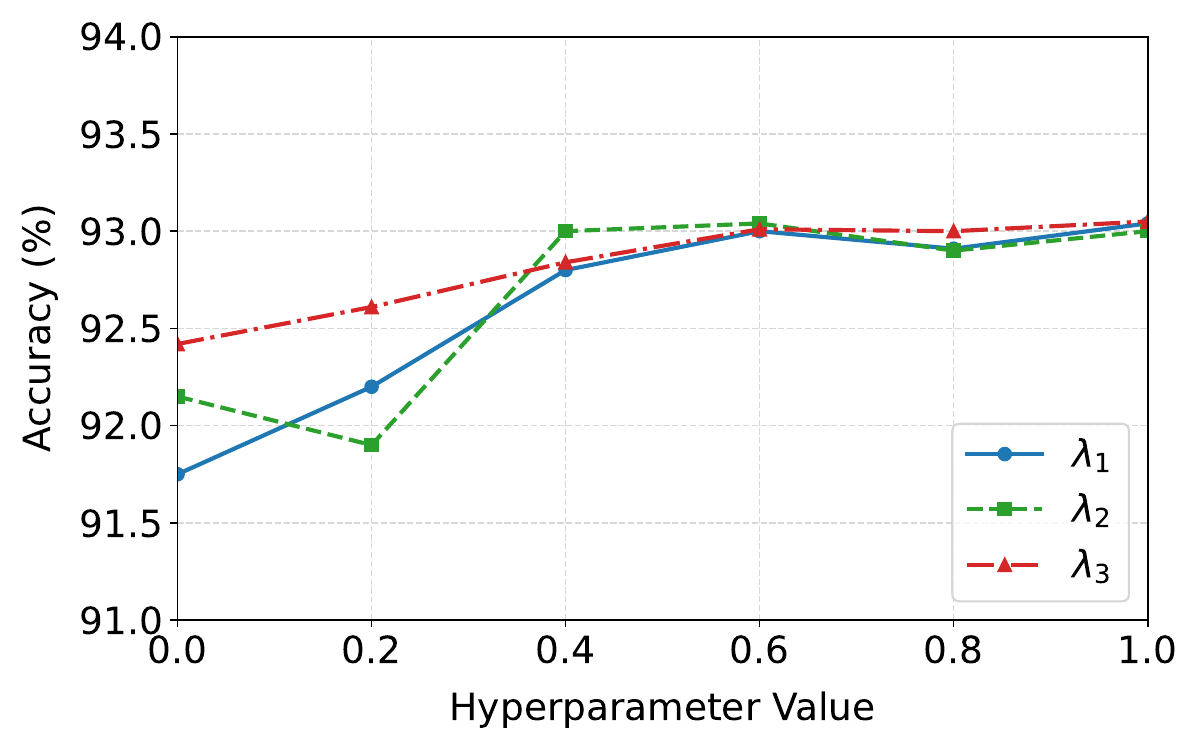}
    \caption{\textbf{Ablation on distillation loss weights.} Top-1 classification accuracy (\%) on CUB-200-2011 obtained by varying the weighting parameters of each distillation loss ($\lambda_1$, $\lambda_2$, and $\lambda_3$) while keeping all other parameter values fixed at 0.5. The accuracy remains stable once the parameters exceed 0.4.}
    \label{fig:ablation_lambda}
\end{figure}

Figure~\ref{fig:ablation_lambda} analyzes the effect of the weighting coefficients associated with the three distillation objectives. Overall, the accuracy rapidly improves as the coefficient increases from 0.0 and then largely saturates once the weight exceeds approximately 0.4. This trend is observed consistently across prediction, patch-level, and attention distillation, indicating that the proposed framework is relatively insensitive to the precise choice of loss weights. %These results suggest that the three distillation objectives provide complementary supervision and do not require extensive hyperparameter tuning to achieve strong performance.

\begin{table}[]
    \centering

    \begin{adjustbox}{width=0.75\linewidth, center}
    \begin{tabular}{cc} \toprule
       Loss term  &  Accuracy\\ \midrule
       $\mathcal{L}_{CE}^{S}$ + $\mathcal{L}_{KD}$  &   91.6$\pm$0.1 \\
       $\mathcal{L}_{CE}^{S}$ + $\mathcal{L}_{patch}$  &  90.8$\pm$0.2 \\
       $\mathcal{L}_{CE}^{S}$ + $\mathcal{L}_{attn}$  &  90.5$\pm$0.1 \\
       $\mathcal{L}_{CE}^{S}$ + $\mathcal{L}_{KD}$ + $\mathcal{L}_{patch}$ & 92.3$\pm$0.1 \\
       $\mathcal{L}_{CE}^{S}$ + $\mathcal{L}_{KD}$ + $\mathcal{L}_{attn}$ & 92.1 $\pm$0.1\\
       $\mathcal{L}_{CE}^{S}$ + $\mathcal{L}_{patch}$ + $\mathcal{L}_{attn}$ & 91.6$\pm$0.1\\    
       $\mathcal{L}_{CE}^{S}$ + $\mathcal{L}_{KD}$ + $\mathcal{L}_{patch}$ + $\mathcal{L}_{attn}$ & \textbf{92.8 $\pm$ 0.2}\\
       \bottomrule
    \end{tabular}
    \end{adjustbox}
    \caption{\textbf{Ablation study of the proposed distillation objectives.} Classification accuracy (\%) obtained using different combinations of distillation losses. All models are trained with fixed weighting parameters for each loss term. The best performance is achieved when all three objectives are combined.}
    \label{tab:distillation_losses}
\end{table}

\textbf{Influence of Text Encoder.}
Table~\ref{tab:text_encoder} shows the impact of different text encoders on FG-LUPI, evaluated using CUB-200-2011. Among the evaluated models, BERT~\cite{devlin2019bert} achieves the highest accuracy, followed closely by RoBERTa~\cite{liu2019roberta} and DeBERTa~\cite{he2020deberta}. The relatively small performance differences among pretrained language models suggest that FG-LUPI is not highly sensitive to the specific choice of text encoder, provided that the textual representations capture meaningful semantic information.

Replacing textual embeddings with random vectors results in a noticeable performance drop, reducing accuracy from 93.0\% to 91.4\%. Interestingly, the random variant still outperforms the vanilla ViT baseline (90.3\%), indicating that part of the improvement stems from the teacher-student distillation framework itself. However, using semantically meaningful textual descriptions yields a further gain of 1.6\% over the random variant, demonstrating that language-derived attribute information provides additional supervisory signals beyond the architectural and distillation benefits alone.

\begin{table}[t]
\centering

\begin{adjustbox}{width=1\linewidth, center}
\begin{tabular}{lccccc}
\toprule
 Encoder & BERT~\cite{devlin2019bert} & RoBERTa~\cite{liu2019roberta} & DeBERTa~\cite{he2020deberta} & CLIP~\cite{radford2021learning} & Random \\
\midrule
Accuracy & \textbf{93.0} & 92.8 & 92.8 & 92.1 & 91.4 \\
\bottomrule
\end{tabular}
\end{adjustbox}
\caption{\textbf{Influence of the text encoder on FG-LUPI performance.} Comparison of different text encoders used to generate attribute representations for the multimodal teacher.}
\label{tab:text_encoder}
\end{table}

\textbf{Effect of EMA and Privileged Information.}
Removing the EMA teacher reduces accuracy from 92.8\% to 91.2\%, suggesting that temporally smoothed teacher representations provide more stable distillation targets during training. Second, replacing the multimodal teacher with an image-only teacher yields 91.2\% accuracy, indicating that the gains are not solely attributable to the teacher-student distillation framework. Instead, the privileged textual descriptions provide complementary semantic supervision that enables the teacher to learn more discriminative representations. %The performance degradation observed in both settings highlights the importance of combining multimodal privileged information with EMA-based online distillation.

\section{Conclusion}

We introduced FG-LUPI, a framework for fine-grained visual classification that formulates textual attribute descriptions as privileged information under the Learning Using Privileged Information (LUPI) paradigm. FG-LUPI leverages textual attribute descriptions during training to provide complementary semantic supervision through a multimodal teacher. This enables the student to benefit from textual attribute information while maintaining standard image-only inference at test time. Experiments across multiple fine-grained recognition benchmarks demonstrate consistent improvements over strong ViT-based baselines and competitive methods. In addition, FG-LUPI achieves lower Expected Calibration Error than the vanilla ViT baseline across all six evaluated benchmarks, suggesting that the benefits of privileged textual supervision extend beyond classification accuracy to empirical confidence calibration. Qualitative analyses further show that the student focuses on meaningful and discriminative image regions, indicating that knowledge acquired from textual supervision can be effectively transferred to an image-only model. Future work will investigate how different forms of privileged information and training objectives influence both predictive performance and confidence calibration in fine-grained recognition.
{
    \small
    \bibliographystyle{ieeenat_fullname}
    \bibliography{main}
}

\clearpage
\appendix

\section{Dataset}\label{app:dataset}

\subsection{Dataset Description}
We evaluate FG-LUPI on six widely used fine-grained visual classification benchmarks spanning multiple domains. 
\begin{itemize}
\item  \textbf{CUB-200-2011}~\cite{wah2011caltech} contains 11,788 bird images from 200 species. The train data
consists of 5,994 images and the test data consists of 5,794 images.
\item \textbf{NABirds}~\cite{van2015building} is a larger bird classification dataset consisting of 48,562 images from 555 North American bird categories. The train set contains 23,929 images and the test set contains 24,633 images.
\item \textbf{Stanford Dogs}~\cite{khosla2011novel} contains 20,580 images covering 120 dog breeds with 12,000 training images and 8,580 test images.
\item \textbf{FGVC-Aircraft}~\cite{maji2013fine} comprises 10,000 aircraft images from 100 aircraft variants, with 6,667 training images and 3,333 test images.
\item \textbf{Stanford Cars}~\cite{krause2013collecting} includes 16,185 images from 196 car categories, with 8,144 training images and 8,041 test images.
\item \textbf{iNaturalist 2021}~\cite{inaturalist21} is a large-scale biodiversity benchmark spanning 10,000 species categories. We conduct experiments on the iNaturalist 2021 Mini subset, using 500,000 training images and 500,000 test images.
\end{itemize}

\subsection{Privileged Information}

For CUB-200-2011, we use the human-authored textual descriptions provided by Reed et al.~\cite{reedLearningDeepRepresentations2016} as privileged information. Each image in the dataset is associated with ten descriptions collected from different Amazon Mechanical Turk workers. During training, we randomly sample one description per image.

For the remaining datasets, human-authored attribute descriptions are unavailable. We therefore generate privileged textual descriptions using multimodal large language models. Specifically, we use Gemma 4 E4B~\cite{google2026gemma4} to generate concise attribute-oriented descriptions for each image. The prompt used for FGVCAircraft, Stanford Cars, and NABirds is shown below, where \lstinline{super_category} is set to `aircraft'', `car'', or ``bird'', respectively.

\begin{lstlisting}[mathescape=true]
You are given a single image of a {super_category}. In ONE concise sentence ($\leq$25 words), describe ONLY the structural and morphological features (body shape, form, proportion, arrangement of parts) that are characteristic of this specific {super_category} type or class. Focus on: shape of the {super_category}, proportions, distinctive structural elements, or anatomical features. Ignore: individual paint schemes, markings, damage, or environmental context. Do NOT name the specific type, species, or breed. Be concise and factual.
\end{lstlisting}

Due to the scale of iNaturalist 2021, instead of image-level caption generation, we generate a single class-level description for each category. For every class, three randomly selected training images are provided to Gemma 3 27B~\cite{teamGemma3Technical2025}, which synthesizes a representative visual description capturing the characteristic appearance of that class. The prompt is shown below.

\begin{lstlisting}
You are given 3 images of a single species. In about 50 words, describe the visual features that are visible in the images and that are characteristic of this class and would help distinguish it from other species in the same supercategory. Focus on color, shape, pattern, texture or distinctive markings. Do NOT name the species/class or add dataset/file metadata. Avoid speculation and hedging (no 'maybe', 'possibly'). Deliver one concise paragraph (~50 words).
\end{lstlisting}

All generated descriptions are used exclusively as privileged information during training and are not required at inference time.

\section{Evaluation Metrics}

We evaluate classification performance using Top-1 Accuracy and Expected Calibration Error (ECE).

\paragraph{Top-1 Accuracy.}
Top-1 Accuracy measures the proportion of test samples for which the predicted class with the highest confidence matches the ground-truth label. Given a test set of size $N$, Top-1 Accuracy is computed as

\begin{equation}
\text{Accuracy}
=
\frac{1}{N}
\sum_{i=1}^{N}
\mathbf{1}
\left(
\arg\max_{c} p_{i,c}
=
y_i
\right),
\end{equation}

where $p_{i,c}$ denotes the predicted probability for class $c$, $y_i$ is the ground-truth label, and $\mathbf{1}(\cdot)$ is the indicator function.

\paragraph{Expected Calibration Error (ECE).}
While accuracy measures predictive correctness, it does not assess whether the model's confidence estimates are reliable. Expected Calibration Error (ECE) quantifies the discrepancy between prediction confidence and empirical accuracy. Following standard practice, predictions are partitioned into $M$ confidence bins. Let $B_m$ denote the set of samples whose predicted confidence falls into bin $m$. ECE is computed as

\begin{equation}
\text{ECE}
=
\sum_{m=1}^{M}
\frac{|B_m|}{N}
\left|
\mathrm{acc}(B_m)
-
\mathrm{conf}(B_m)
\right|,
\end{equation}

where $\mathrm{acc}(B_m)$ is the average accuracy of samples in bin $m$, $\mathrm{conf}(B_m)$ is the average predicted confidence in the same bin, and $|B_m|$ denotes the number of samples assigned to the bin. Lower ECE values indicate better calibration, with an ECE of zero corresponding to perfectly calibrated predictions.

\section{Additional Results} \label{app:additional_results}

\begin{figure}
\begin{tabular}{@{}ccc@{}} 
Original  & Vanilla ViT & FG-LUPI \\
\includegraphics[width=0.3\linewidth]{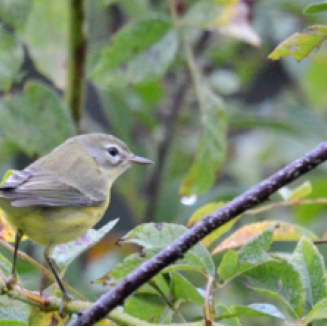} &
\includegraphics[width=0.3\linewidth]{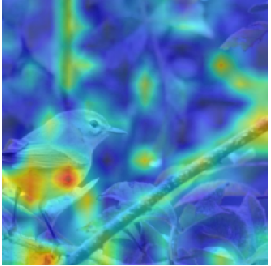} &
\includegraphics[width=0.3\linewidth]{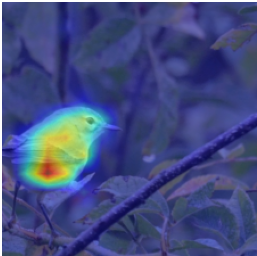}   \\
\includegraphics[width=0.3\linewidth]{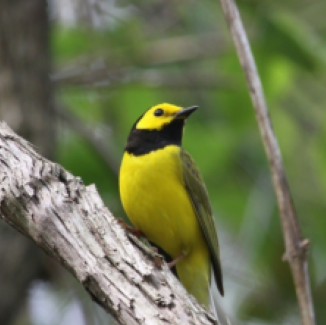} &
\includegraphics[width=0.3\linewidth]{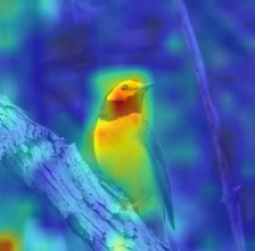} &
\includegraphics[width=0.3\linewidth]{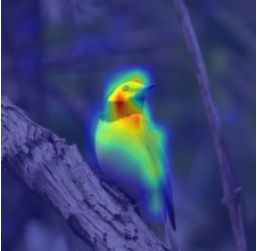}\\
\includegraphics[width=0.3\linewidth]{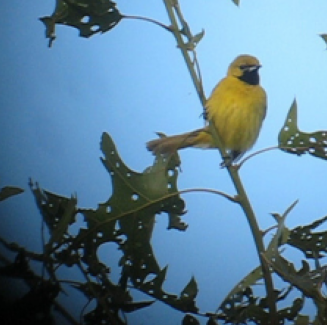} &
\includegraphics[width=0.3\linewidth]{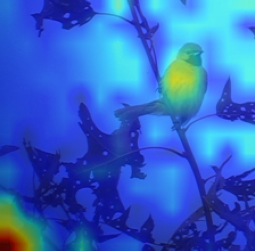} &
\includegraphics[width=0.3\linewidth]{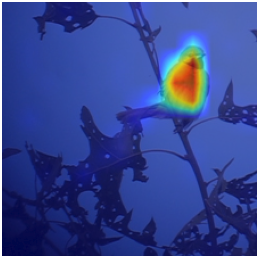}\\

\end{tabular}

\caption{\textbf{Cross-dataset attention generalization.} Attention maps produced by a vanilla ViT and FG-LUPI on NABirds images using models trained only on CUB-200-2011. Despite never being trained on NABirds, FG-LUPI exhibits more concentrated attention on bird regions while reducing attention to surrounding background. In contrast, the vanilla ViT produces more diffuse attention. These qualitative results suggest that knowledge learned under textual attribute supervision transfers beyond the training dataset.}
\label{fig:cross_dataset_attn}
\end{figure}

\begin{figure}
\centering

\setlength{\tabcolsep}{2pt}
\renewcommand{\arraystretch}{0.9}

\begin{tabular}{@{}ccc@{}} 
  & black eyes & yellow breast \\
\includegraphics[width=0.3\linewidth]{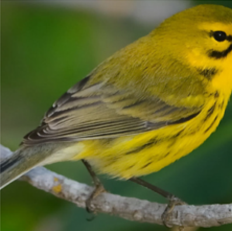} &
\includegraphics[width=0.3\linewidth]{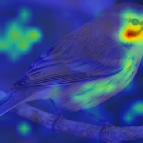} &
\includegraphics[width=0.3\linewidth]{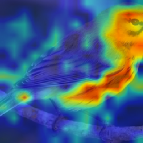}   \\
  & orange wing & black neck \\
\includegraphics[width=0.3\linewidth]{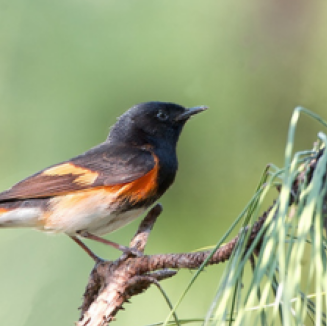} &
\includegraphics[width=0.3\linewidth]{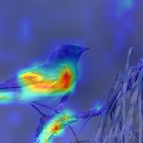} &
\includegraphics[width=0.3\linewidth]{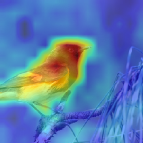}\\
  & white belly & yellow beak \\
\includegraphics[width=0.3\linewidth]{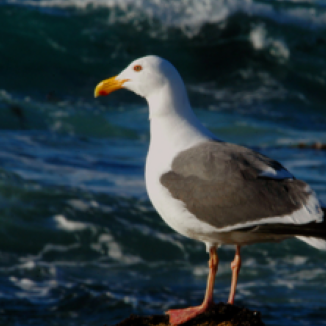} &
\includegraphics[width=0.3\linewidth]{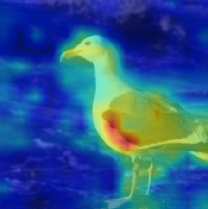} &
\includegraphics[width=0.3\linewidth]{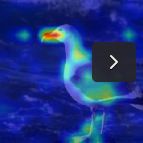}\\
\end{tabular}

\caption{\textbf{Text-conditioned attention visualization of the multimodal teacher network.} Given different attribute descriptions, we visualize the corresponding cross-attention maps for the same input image. Different attribute descriptions produce distinct attention patterns, indicating that the teacher learns attribute-dependent interactions between textual and visual representations.}
\label{fig:text_grounding}
\end{figure}

\begin{figure*}[t]
\centering
\setlength{\tabcolsep}{1pt} % or 0pt

\begin{tabular}{@{}cccccc@{}}
Original & Vanilla ViT & FG-LUPI & Original & Vanilla ViT & FG-LUPI\\

\includegraphics[width=0.16\linewidth, height = 2.6cm]{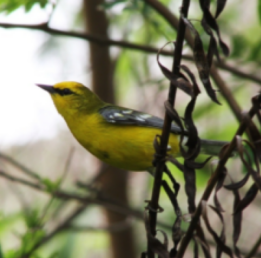} &
\includegraphics[width=0.16\linewidth, height = 2.6cm]{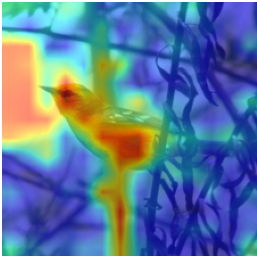} &
\includegraphics[width=0.16\linewidth, height = 2.6cm]{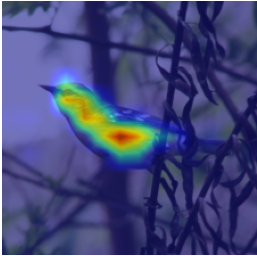} &

\includegraphics[width=0.16\linewidth, height = 2.6cm]{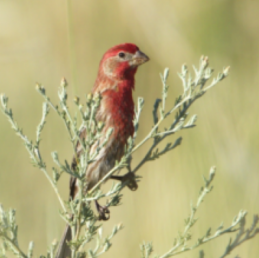} &
\includegraphics[width=0.16\linewidth, height = 2.6cm]{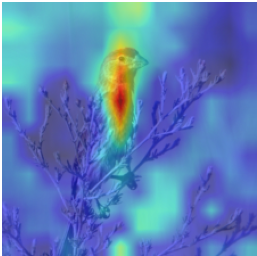} &
\includegraphics[width=0.16\linewidth, height = 2.6cm]{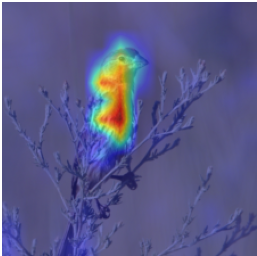} \\

\includegraphics[width=0.16\linewidth, height = 2.6cm]{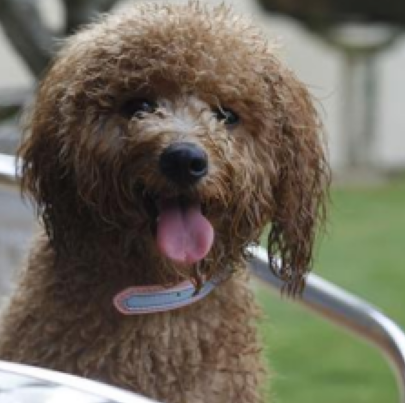} &
\includegraphics[width=0.16\linewidth, height = 2.6cm]{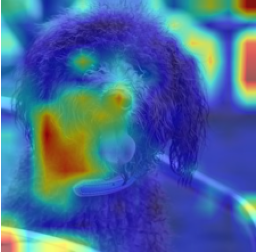} &
\includegraphics[width=0.16\linewidth, height = 2.6cm]{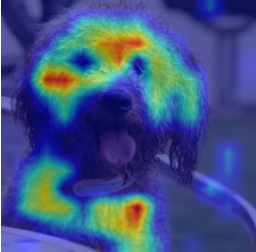} &

\includegraphics[width=0.16\linewidth, height = 2.6cm]{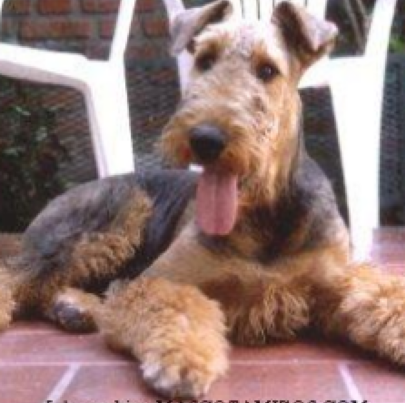} &
\includegraphics[width=0.16\linewidth, height = 2.6cm]{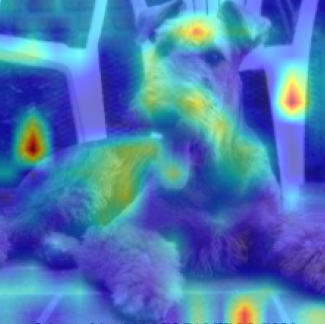} &
\includegraphics[width=0.16\linewidth, height = 2.6cm]{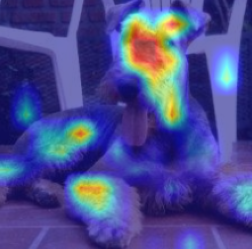} \\

\includegraphics[width=0.16\linewidth, height = 2.6cm]{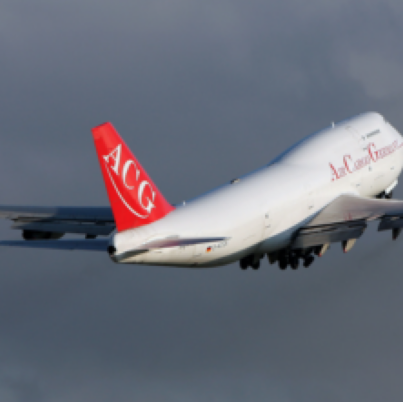} &
\includegraphics[width=0.16\linewidth, height = 2.6cm]{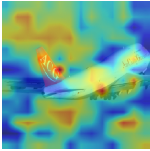} &
\includegraphics[width=0.16\linewidth, height = 2.6cm]{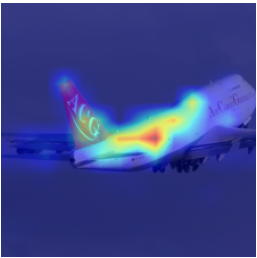} &

\includegraphics[width=0.16\linewidth, height = 2.6cm]{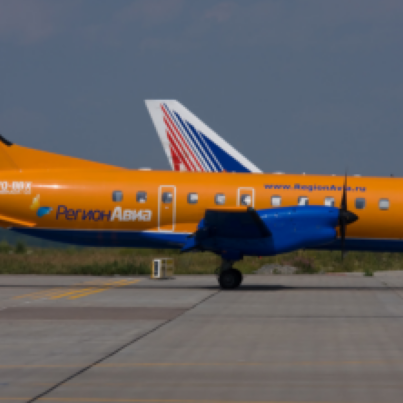} &
\includegraphics[width=0.16\linewidth, height = 2.6cm]{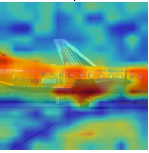} &
\includegraphics[width=0.16\linewidth, height = 2.6cm]{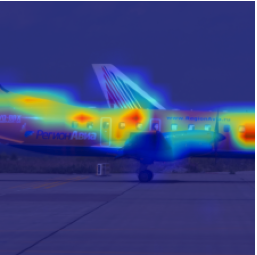} \\

\includegraphics[width=0.16\linewidth, height = 2.6cm]{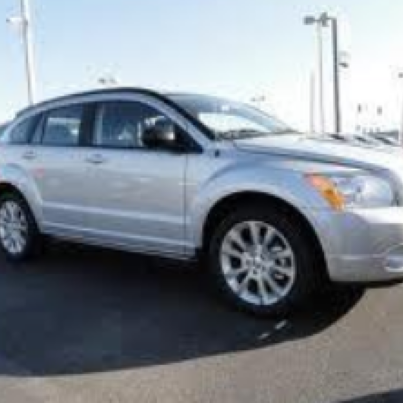} &
\includegraphics[width=0.16\linewidth, height = 2.6cm]{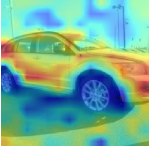} &
\includegraphics[width=0.16\linewidth, height = 2.6cm]{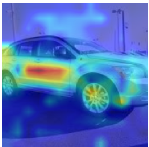} &

\includegraphics[width=0.16\linewidth, height = 2.6cm]{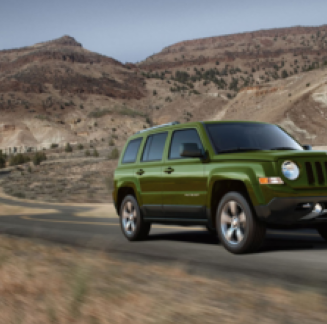} &
\includegraphics[width=0.16\linewidth, height = 2.6cm]{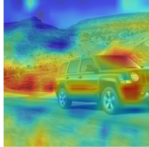} &
\includegraphics[width=0.16\linewidth, height = 2.6cm]{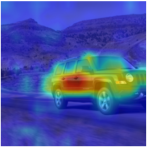} \\

\end{tabular}

\caption{\textbf{Qualitative comparison of attention maps.} Attention map visualization on NABirds, Stanford dogs, FGVC Aircraft and Stanford Cars for vanilla ViT and the FG-LUPI student model. Compared to the vanilla model, FG-LUPI exhibits more concentrated attention on the object and reduced attention to surrounding background regions.}

\label{fig:vit_vs_ours_add}
\end{figure*}

\textbf{Cross-dataset attention generalization.} 
To investigate whether the learned localization behavior generalizes beyond the training distribution, we visualize attention maps on NABirds using models trained exclusively on CUB-200-2011. As shown in Fig.~\ref{fig:cross_dataset_attn}, FG-LUPI concentrates on the bird and its discriminative anatomical regions despite never observing NABirds images during training. In contrast, the vanilla ViT frequently allocates attention to surrounding context such as vegetation, water, and background structures. These results suggest that privileged attribute supervision encourages the model to learn transferable part-based localization cues rather than dataset-specific correlations. The learned attention patterns therefore exhibit stronger robustness to background clutter and generalize effectively to unseen fine-grained bird categories.

\textbf{Text-conditioned attention visualization. } To further analyze the behavior of the multimodal teacher, we visualize the cross-attention maps conditioned on different textual attribute queries on CUB-200-2011. As shown in Fig.~\ref{fig:text_grounding}, the same input image produces distinct attention responses depending on the queried attribute. For instance, when prompted with ``black eyes'', the model focuses on the head region, while ``yellow breast'' shifts attention to the chest area, and ``orange wing'' highlights the wing region. 
We also observe that certain attributes exhibit spatial overlap or ambiguity in their visual grounding. For example, ``black neck'' may partially activate regions around both the neck and adjacent wing area, while ``white belly'' can extend slightly toward nearby regions. Since attribute phrases such as ``white belly'' or ``black neck'' are encoded as separate tokens, the model aggregates token-level semantics rather than treating each attribute as an atomic region descriptor. As a result, the induced attention may reflect a soft combination of multiple semantic cues, leading to partially overlapping activation across adjacent body parts.

\textbf{Additional Qualitative Results.} Additional qualitative results are shown in Fig.~\ref{fig:vit_vs_ours_add} across NABirds, Stanford Dogs, FGVC Aircraft, and Stanford Cars. Compared to a vanilla ViT, the FG-LUPI student exhibits more concentrated attention on the foreground object while reducing attention to surrounding background regions. This behavior is observed across diverse object categories, including birds, dogs, aircraft, and vehicles, suggesting that the knowledge transferred from the multimodal teacher encourages the student to focus more on relevant object regions. %These observations provide qualitative evidence that the student retains aspects of the teacher's attention structure after distillation.

\begin{table*}[h]

    \begin{adjustbox}{width=0.94\linewidth, center}
    \centering
    \begin{tabular}{ccccccccccccc} \toprule
      \multirow{2}{*}{Method} & \multirow{2}{*}{Backbone} & \multirow{2}{*}{Parameters} & \multicolumn{5}{c}{Zero-shot} & \multicolumn{5}{c}{Fine-tuned} \\
      \cmidrule(lr){4-8} \cmidrule(lr){9-13}
      & & & CUB & Dogs & NABirds & Aircraft & Cars & CUB & Dogs & NABirds & Aircraft & Cars\\ \midrule
      \multirow{3}{*}{CLIP} & ViT-B/32 & 151.3M & 53.3 & 43.1 & 36.0 & 18.5 & 50.7 & 71.6 & 70.0 & 60.8 & 69.6 & 82.5\\
      & ViT-B/16 & 149.6M & 55.4 & 47.5 & 40.0 & 24.2 & 58.2 & 78.7 & 74.8 & 75.7 & 78.2 & 87.4\\
      & ViT-L/14 & 427.6M & 62.0 & 60.7 & \underline{53.4} & \textbf{35.3} & 73.3 & 85.4 & 88.3 & 85.4 & 89.9 & 88.5\\ \midrule
      SigLIP & ViT-B/16 & 203.2M & \textbf{71.1} & \textbf{76.2} & 50.8 & 21.3 & \textbf{95.1} & 81.6 & 86.1 & 77.3 & 87.4 & \textbf{96.0}\\ \midrule
      \multirow{2}{*}{CoCa} & ViT-B/32 & 255.5M & 51.6 & 44.9 & 40.7 & 12.4 & 76.5 & 72.8 & 73.8  & 67.4 & 66.0 & 91.7\\
       & ViT-L/14 & 644.0M & \underline{70.2} & \underline{67.8} & \textbf{58.9} & \underline{24.6} & \underline{92.2} & 86.5 & 85.5 & 86.0 & 90.1 & \underline{95.7}\\ \midrule
      \multirow{2}{*}{BLIP} & ViT-B/16 & 223.7M & 17.8 & 34.5 & 12.8 & 5.6 &  66.2 & 55.3 & 66.6 & 72.6 & 37.8 & 84.9\\ 
      & ViT-L/16 & 446.1M & 25.5 & 37.9 & 20.3 & 6.7 & 70.0 & 66.0 & 71.9 & 76.3 & 34.7 & 85.8\\ 
      \midrule
      \multirow{1}{*}{VLM2Vec} & Qwen2-VL-2B & 2B & 2.7 & 3.8 &  1.7 & 2.5 & 1.2 & -- & -- & -- & -- & -- \\ \midrule
      FG-LUPI Teacher & ViT-B/16 & 216.5 M & -- & -- & -- & -- & -- & \underline{91.4} & \underline{93.0} & \underline{91.8} & \underline{92.6} & 87.0 \\
      FG-LUPI Student & ViT-B/16 & 86.7 M & -- & -- & -- & -- & -- & \textbf{92.8} & \textbf{95.1} & \textbf{92.3} & \textbf{94.0} & 89.2 \\ \bottomrule

    \end{tabular}
\end{adjustbox}
    \caption{\textbf{Top-1 classification accuracy (\%) on vision-language models} under zero-shot and supervised fine-tuning settings, alongside the FG-LUPI Teacher and image-only Student. The best and second-best fine-tuned results are shown in bold and underlined, respectively.}

    \label{tab:results_vlm}
\end{table*}

\textbf{Impact of Text Source.}
We further investigate whether the effectiveness of FG-LUPI depends on the source of the textual descriptions. On CUB-200, we compare human-provided attribute descriptions with LLM-generated descriptions using Gemma 4 EB. Using human text, FG-LUPI achieves $92.8 \pm 0.2\%$ accuracy, while LLM-generated descriptions achieve $92.6 \pm 0.1\%$. The comparable performance suggests that FG-LUPI is relatively robust to the source of textual supervision, with semantically relevant descriptions providing effective guidance for visual representation learning regardless of whether they are human- or LLM-generated.

\textbf{Comparison with Vision-Language Models.} Table~\ref{tab:results_vlm} compares FG-LUPI with representative VLMs, including CLIP~\cite{radford2021learning}, SigLIP~\cite{zhai2023sigmoid}, CoCa~\cite{yu2022coca}, BLIP~\cite{li2022blip}, and VLM2Vec~\cite{jiang2024vlm2vec}. 
For VLM2Vec, we report zero-shot performance using the Qwen2-VL-2B variant. While VLMs such as CLIP, BLIP, SigLIP, and CoCa explicitly optimize image-text alignment in ways that support direct zero-shot classification, VLM2Vec is designed as a general multimodal embedding model trained across a broader range of embedding tasks. Its substantially lower zero-shot performance therefore suggests that its pretrained embedding space is less directly aligned with the class-level and fine-grained distinctions required for FGVC.
While existing VLMs can achieve strong performance after fine-tuning, FG-LUPI obtains competitive or superior results across the evaluated fine-grained datasets. FG-LUPI student achieves the best performance on CUB, Dogs, NABirds, and FGVCAircraft, despite having only 86.7M parameters compared to substantially larger VLMs. These results suggest that privileged text supervision during training can effectively transfer fine-grained semantic guidance to a lightweight image-only model, yielding strong recognition performance without requiring text inputs at inference.

\textbf{Computational Overhead Analysis.} To quantify the training overhead of our framework at $448 \times 448$ resolution and a batch size of 1, we benchmarked the training latency against a standard ViT-B/16 baseline. Our combined framework (Multimodal Teacher + Student) requires $13.39 \pm 0.92\text{ ms}$, while the standard ViT-B/16 baseline requires $7.02 \pm 0.29\text{ ms}$. This corresponds to an approximately $1.9\times$ increase in training latency. The additional cost primarily arises from the multimodal Teacher, including text-image fusion and the associated forward and backward computations. Importantly, this overhead is incurred only during training. At deployment, the Teacher and text-processing components are discarded, leaving the image-only Student with 86.7M parameters and no additional inference-time computation compared to the corresponding ViT-B/16 image-only model.

\section{Limitations and Future Work}
\label{sec:limitations}

Despite the consistent improvements across the evaluated benchmarks, FG-LUPI has several limitations. First, the effectiveness of the proposed training-time textual supervision depends on the specificity and quality of the available descriptions. This is particularly relevant for fine-grained categories such as Stanford Cars, where visually similar classes may differ in highly localized and model-specific details that are not always captured by generated descriptions. Second, while the CUB descriptions are human-authored and constitute externally provided privileged information, descriptions for the other datasets are synthetically generated from the training images using Gemma. These descriptions should therefore be interpreted as training-time privileged semantic supervision rather than independent sources of visual information. Their purpose is to expose semantic structure that can guide the multimodal teacher and subsequently be transferred to the image-only student.
The additional multimodal teacher introduces computational cost during training, although the textual modality and teacher are discarded at inference and the deployed FG-LUPI student requires only an image. 

Future work will investigate richer and more reliable sources of privileged semantic supervision, including fine-grained human annotations and automatically learned attribute representations. It would also be valuable to study whether the proposed framework can exploit other forms of training-only information, as well as extend the approach to settings where textual supervision varies in quality.

\end{document}